\documentclass[letterpaper, 10 pt, conference]{ieeeconf}  

\IEEEoverridecommandlockouts                              

\usepackage{graphics} 
\usepackage{epsfig} 
\usepackage{mathptmx} 
\usepackage{times} 
\usepackage{amsmath} 
\usepackage{amssymb}  
\usepackage{color,soul}
\usepackage{multirow}
\usepackage{color}
\title{\LARGE \bf
Large Scale Visual Place Recognition with Sub-Linear Storage Growth
}

\author{Huu Le and Michael Milford
\thanks{The authors are with the School of Electrical Engineering and Computer Science,
        Queensland University of Technology, Brisbane, Australia. MM is also with the Australian Centre for Robotic Vision. This work was supported by an Asian Office of Aerospace Research and Development Grant FA2386-16-1-4027 and an ARC Future Fellowship FT140101229 to MM.  Email: \{huu.le, michael.milford\}@qut.edu.au
}%
}

\begin{document}

\maketitle
\thispagestyle{empty}
\pagestyle{empty}

\begin{abstract}

Robotic and animal mapping systems share many of the same objectives and challenges, but differ in one key aspect: where much of the research in robotic mapping has focused on solving the data association problem, the grid cell neurons underlying maps in the mammalian brain appear to intentionally break data association by encoding many locations with a single grid cell neuron. One potential benefit of this intentional aliasing is both sub-linear map storage \textit{and} computational requirements growth with environment size, which we demonstrated in a previous proof-of-concept study that detected and encoded mutually complementary co-prime pattern frequencies in the visual map data. In this research, we solve several of the key theoretical and practical limitations of that prototype model and achieve significantly better sub-linear storage growth, a factor reduction in storage requirements per map location, scalability to large datasets on standard compute equipment and improved robustness to environments with visually challenging appearance change. These improvements are achieved through several innovations including a flexible user-driven choice mechanism for the periodic patterns underlying the new encoding method, a parallelized chunking technique that splits the map into sub-sections processed in parallel and a novel feature selection approach that selects only the image information most relevant to the encoded temporal patterns. We evaluate our techniques on two large benchmark datasets with comparison to the previous state-of-the-art system, as well as providing detailed analysis of system performance with respect to parameters such as required precision performance and the number of cyclic patterns encoded. {\color{white} \cite{yu2018rhythmic} }

\end{abstract}

\section{Introduction}
Recent rapid advances in camera technology, computing power and application areas in robotics and autonomous vehicles have resulted in a surge in Visual Place Recognition research~\cite{lowry2016visual,lynen2015get, cadena2016past}, a key enabling component of many localization and SLAM systems. Much recent research has focused on the key challenges of deploying these techniques reliably on mobile robots, drones and autonomous vehicles, including being robust to varying environmental conditions and camera viewpoints. Much of the current research has focused primarily on improving standard metrics such as precision and recall on benchmark datasets, while storage and computational requirements, especially with respect to their \textit{scalability}, have not been as thoroughly investigated. As the size of the deployment area for a drone, autonomous vehicle or mobile robot increases, so too does the storage and compute size if performance is to be maintained: this increase is typically at least linear with the size of the environment in current approaches. While the absolute capability of compute and storage technology continues to improve, efficient algorithmic implementations are always desirable~\cite{bojarski2016end}, as they provide the option to encode more or richer data, perform more extensive computation and deploy on cheaper, more compact and lower power consumption computational hardware.

One unorthodox but promising area of inspiration for efficient visual place recognition algorithms is biology. In 2004 neuroscientists discovered a new type of spatial mapping cell called a grid cell~\cite{fyhn2004spatial}, which appeared to defy robotic mapping convention. Each grid cell fired when the rat was at any one of a near unlimited number of physical locations at the vertices of a regular tesselating triangular grid over the environment. Multiple map scales were encoded in parallel, leading neuroscientists and mathematicians to posit a range of theories~\cite{burak2009accurate,sreenivasan2011grid} around their function including noise rejection and efficient computation. One of the first studies~\cite{yu2018rhythmic} to develop a robot-deployable version of these theories proposed a novel encoding scheme that mimicked the deliberate aliasing of place to map associations, demonstrating for the first time sub-linear map storage growth with environment size.

\begin{figure}[!]
    \centering
    \includegraphics[width = 0.99\columnwidth]{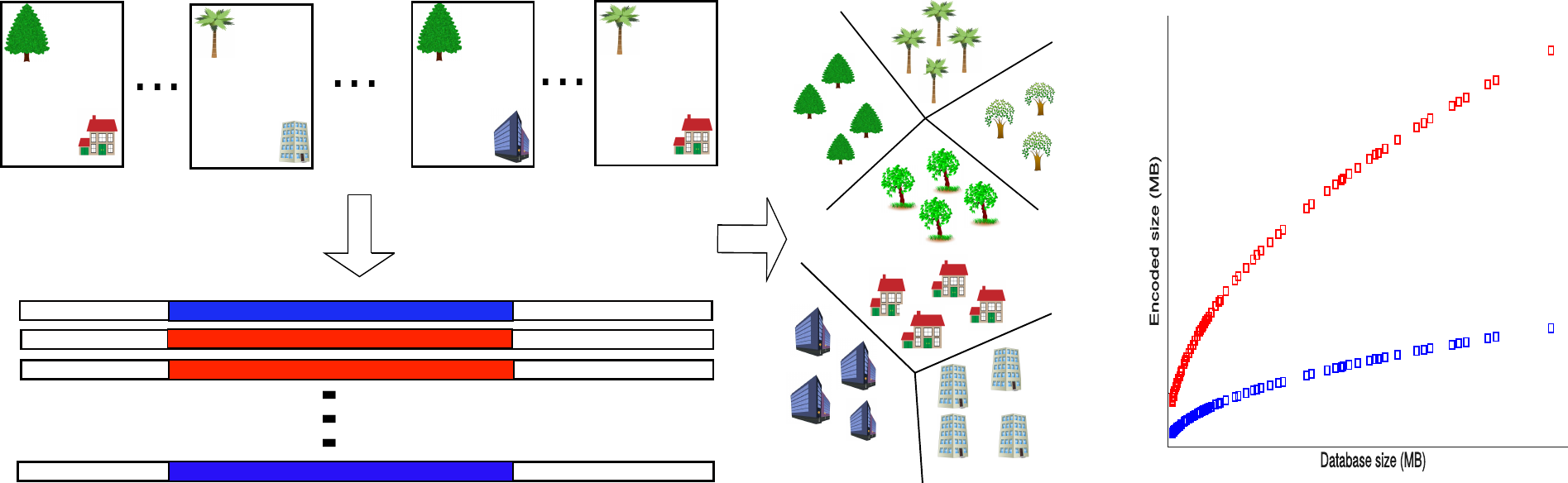}
    \caption{Overview of our sub-linear encoding system. Left: Each frame in the sequence is represented by a feature vector. The pattern learning mechanism in our system automatically learns the most relevant information that represents the multiple periodic patterns detected in the data, which are then used to train the encoding system. Right: Our new system (blue) achieves significant improvements in both sub-linear storage scaling and absolute storage requirements over the previous state-of-the-art method proposed in~\cite{yu2018rhythmic} (red) while maintaining recognition performance and improving robustness to changing conditions and customizability to user requirements. }
    \label{fig:overview}
\end{figure}

In this new research we present a range of novel theoretical and practical contributions that address the major shortcomings of that prototype study including limited sub-linearity, a reliance on finding fixed co-prime frequency patterns, limitations in scaling to larger datasets and robustness to changing environmental conditions. To achieve these improvements, we make a number of contributions as follows:

\begin{itemize}
        \item We present a new encoding algorithm that allows the pattern periods underlying the encoding method to be chosen arbitrarily, giving the user more flexbility in fine-tuning the required performance-storage growth tradeoff that best matches the application requirements
        \item To efficiently encode only the image information directly relevant to the detected period patterns in the data, we introduce a novel feature selection approach, leading to a drastic reduction in the absolute amount of storage required while maintaing or improving on the accuracy achieved by~\cite{yu2018rhythmic}. This reduction in absolute storage size also improves the query time response.
        \item To handle large datasets, we provide a sequence splitting scheme that partition the data into multiple sub-sequences that can be trained in parallel while maintaining sub-linearity storage growth, reducing the training time while still maintaining overall accuracy performance.
        \item To further improve robustness to changing environmental conditions, we update the feature front-end with a just released LostX~\cite{garg2018lost} feature type.
        \item We present a new GPU and multi-core based implementation that maximizes usage of current modern computational hardware, and provide our implementation as open source code at https://github.com/intellhave/SublinearEncoding
\end{itemize}

Together these improvements enable us to train and benchmark our system on large datasets containing over 80,000 frames with comparison to the previous state-of-the-art system, achieving significantly improved sub-linear storage growth and an equivalent reduction in computational requirements while maintaining accuracy performance.

The rest of the paper proceeds as follows. In Section~\ref{sec:background}, we review several representation learning and database encoding schemes that are related to our work. Section~\ref{sec:approach} describes our database sub-linear encoding method, followed by the sequence chunking approach to handle large scale datasets. The experiments and detailed analysis of our proposed method on different real-world datasets are provided in Section~\ref{sec:experiments}. Finally, in Section~\ref{sec:conclusion}, we discuss possible directions for future work and conclude the paper.

\section{Background}
\label{sec:background}
In this section, we briefly review some popular techniques for the task of visual place recognition in SLAM, followed by the motivations behind our work.

Early methods for visual place recognition rely on the use of several feature indexing techniques such as bag-of-words~\cite{filliat2007visual,sivic2003video} or vocabulary tree~\cite{nister2006scalable}. For each frame in the training database, a set of local features (for instance, SIFT) are extracted. The collection of all extracted features are then quantized into visual words~\cite{gersho2012vector}. Popular weighting schemes such as ``term frequency - inverse document frequency" (TF-IDF) are used to rank the visual words and each image is stored in the database as a vector of word frequencies~\cite{sivic2003video}. The vocabulary tree indexing method~\cite{nister2006scalable} made a slight modification to the quantization process, in which the local features are quantized hierarchically into a vocabulary tree. This approach led to an efficient method with higher accuracy compared to the conventional bag-of-word approach with faster training and querying time.

In the field of data compression and nearest neighbor search for media retrieval, variants of the vector quantization (VQ) approach~\cite{gersho2012vector}, such as product quantization (PQ)~\cite{jegou2011product,kalantidis2014locally} are often used in many encoding and retrieval applications. VQ learns a dictionary containing $k$ code-words using the K-Means~\cite{macqueen1967some} clustering algorithm. After the dictionary is obtained, each input data is represented by its nearest code-word in the dictionary. Therefore, an input data can be stored using $log_2 k$ bits. When the dictionary size $k$ becomes very large, the use of K-Means with high dimensional data is very computationally expensive. Therefore, PQ decomposes the original feature space into many orthogonal sub-spaces and performs VQ for each individual subspace. Different methods have also been proposed based on PQ~\cite{kalantidis2014locally,jegou2012aggregating,jegou2010aggregating} for image and media retrieval.

Learning binary representation~\cite{wang2018survey,gong2013iterative} is also a popular approach for image retrieval. Instead of using code-word index as proposed by VQ technique, each input data point is embedded into a compact binary domain. The nearest distance between two vectors in the original space can be efficiently using the Hamming distance in the binary space. Different deep-learning based binary hashing techniques have been proposed~\cite{lin2015deep,liu2016deep,he2013k}

Among a large body of work on media retrieval, there has been relatively little attention on the problem of sub-linear storage scaling of the encoded data. This work is developed based on the initial approach proposed in~\cite{yu2018rhythmic}, which proposes a database encoding algorithm that achieved sub-linear scale by detecting and encoding co-prime period patterns in the underlying map data.
\section{Approach}
\label{sec:approach}
Here we describe our new data encoding approach for visual place recognition. As previously mentioned, our method encodes a dataset in such a way that the storage required for the compressed data scales sub-linearly with respect to the growth of the training data. In addition, by utilizing current state-of-the-art results on visual place recognition under changing conditions, our proposed method is able to handle localization tasks where the query scenes undergo day/night or seasonal changes, as will be shown in our experiments. The efficiency in our algorithm is achieved by a new database encoding scheme based on repeated patterns, and a feature selection strategy that only learns the most distinctive information for a specific pattern based on a pre-defined cycle assignment, which will be explained in more detail in the following sections.
\subsection{Database encoding with repeated patterns}
\label{sec:database_encoding}
\begin{figure}
    \centering
    \includegraphics[width=0.99\columnwidth]{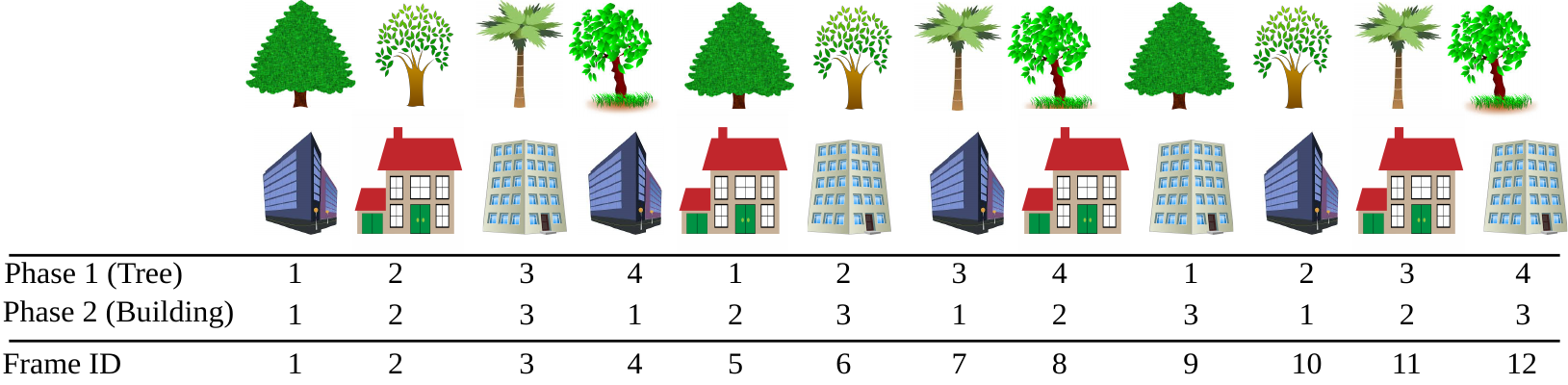}
    \caption{Illustration of place recognition using cyclic patterns.}\vspace{-0.5cm}
    \label{fig:repeated_patterns}
\end{figure}
\begin{figure}
    \centering
    \includegraphics[width=0.99\columnwidth]{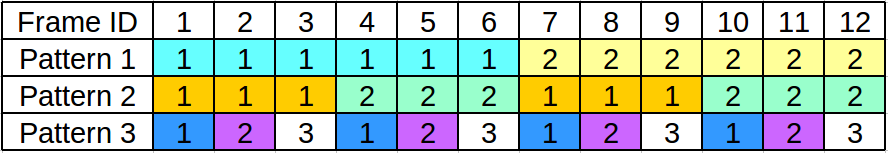}    
    \caption{Example of label assignments to train the phase encoders for $N=12$ frames, encoded by 3 cyclic patterns with cycle lengths of $\tau_1=2$, $\tau_2=2$ and $\tau_3=3$ .} \vspace{-0.5cm}
    \label{fig:phase_assignment}
\end{figure}
Inspired by the work of Yu et al~\cite{yu2018rhythmic} and recent research on navigation neurons in the brain of mammals~\cite{giocomo2007temporal}, the method proposed in this work also takes advantage of multiple repeated visual patterns for the task of visual place recognition. For completeness, Fig.~\ref{fig:repeated_patterns} (which is reproduced from~\cite{yu2018rhythmic}) shows a simple illustration of location identification using two cyclic patterns: trees and buildings. The first cyclic pattern (tree) has cycle length of $\tau_1=4$, while the second pattern (building) repeats after every $\tau_2=3$ frames. These cyclic patterns allow a database with $N=12$ frames to be encoded. During the training process, each cyclic pattern is associated with a \emph{phase encoder}, which is trained to learn the templates corresponding to the phases in that particular cycle. After the phase encoders are successfully trained, given a new scene, its location with respect to the training dataset can be uniquely identified based on the phases decoded from the phase encoders (as illustrated in Fig.~\ref{fig:repeated_patterns}).

The approach mentioned above has been employed in the work of~\cite{yu2018rhythmic}. However, as the phase labels are assigned sequentially for all $k$ cyclic visual patterns,~\cite{yu2018rhythmic} requires the cycle lengths $\tau_1, \dots,\tau_k$ to be co-primes, and $\tau_1\tau_2\dots \tau_k \ge N$ (where $N$ is the number of training frames) in order for a scene to be uniquely identified. As the value of $N$ grows, the large gaps between co-prime numbers can result in unnecessary storage as the product $\tau_1\tau_2\dots \tau_k$ can be much greater than $N$. We address this issue in this work by proposing a new database learning scheme such that the cycle lengths \emph{are not required to be co-primes}. This gives us more flexibility in selecting the desired hyper parameters that minimize the amount of storage required. 

Fig.~\ref{fig:phase_assignment} shows an example of how the labels are assigned to train our phase encoders for 3 cyclic patterns with cycle length of $\tau_1=2$, $\tau_2=2$ and $\tau_3=3$, respectively. For brevity, we assume that $\tau_1, \dots, \tau_k$ are chosen such that $N = \tau_1\tau_2\dots \tau_k$. Observe from Fig.~\ref{fig:phase_assignment} that the labels for the phase encoders are assigned in a hierarchical manner. Particularly, the input data are partitioned into $\tau_1$ groups, labelled from $1$ to $\tau_1$ (second row in Fig.~\ref{fig:phase_assignment}). Each group in the first pattern are then further divided into $\tau_2$ groups, which are then labelled from $1$ to $\tau_2$. This process of partitioning and labelling is continued recursively for all $k$ cyclic patterns.

By employing this hierarchical scheme of assigning the labels to the phase encoders, the cycle lengths $\tau_1,\dots,\tau_k$ are not required to be co-primes. In fact, it can be seen from Fig.~\ref{fig:phase_assignment} that, for each scene $s_i$, the labels $l^i_1,\dots,l^i_k$ of the phase encoders form an element of the $k-$permutation of $\{1,\dots,\hat{\tau}\}$, where $\hat{\tau} = \max(\tau_1,\dots,\tau_k)$. Therefore, it can be assured that each scene in the database can be uniquely identified by the set of phase encoder labels $l^i_1,\dots,l^i_k$. Formally the index $i$ of a scene can be computed from $l_1,\dots,l_k$ as


\begin{equation}\label{eq:scene_index}
    i = (l^i_1 - 1)\frac{N}{\tau_1} + (l^i_2 - 1)\frac{N}{\tau_1 \tau_2} + \dots +(l^i_{k-1} - 1)\frac{N}{\tau_1 \tau_2\dots \tau_k} + l^i_k.
\end{equation}

After all the phase encoders have been trained, to localize a new scene $\hat{s}$, it is passed into all $k$ phase encoders to obtain the set of phase prediction $\hat{l}_1,\dots,\hat{l}_k$. Then, the index of the query scene with respect to the training dataset is computed using~\eqref{eq:scene_index}. As the index can be computed directly, our query time is also faster than~\cite{yu2018rhythmic}, as~\cite{yu2018rhythmic} requires an addition step of set intersection for index computation. Due to the sub-linear storage requirement, similar to~\cite{yu2018rhythmic}, we also choose our phase encoders to be linear SVMs.



\subsection{Pattern learning}
\label{sec:pattern_learning}
\begin{figure}
    \centering
    \includegraphics[width = 0.9\columnwidth]{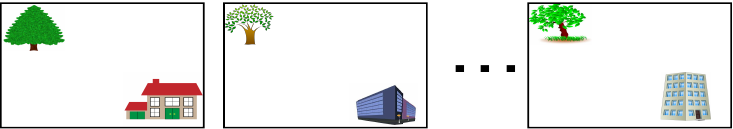}
    \caption{Example of a frame sequence with two cyclic patterns: tree and building, where trees only appear at the top-left corner of the scenes, while buildings only appear at the bottom-right corner of the frames.}
    \label{fig:pattern_learning}
\end{figure}

\begin{figure}
    \centering
    \includegraphics[width = 0.8\columnwidth]{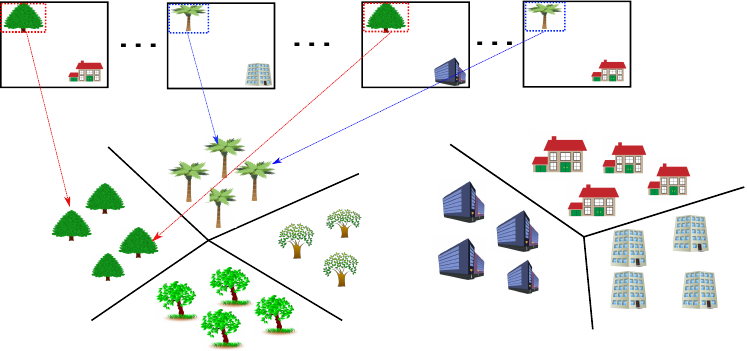}
    \caption{Illustration of pattern learning. Pattern learning selects the regions of the images such that regions containing the same template are grouped together. } \vspace{-0.5cm}
    \label{fig:pattern_group}
\end{figure}
In addition to the new phase encoding scheme as discussed in the previous section, a key improvement of our work compared to~\cite{yu2018rhythmic} lies in our ability to explicitly learn the patterns from the training data. This pattern learning mechanism is developed based on the observation that, for a specific pattern, only a small region of the image contains the object. See Fig.~\ref{fig:pattern_learning} for an illustration of a frame sequence with two cyclic patterns, in which the objects of the first pattern (trees) only appear at the top-left corner of the images, while the objects representing the second pattern (buildings) only appear at the bottom-right corner. By explicitly extracting only the regions containing the objects that represent the cyclic patterns to train the encoders, our method can significantly reduce the absolute amount of storage required, while the accuracy of the pattern phase encoders can be improved, as redundant information is eliminated during the training process. 

The mechanism of learning the cyclic patterns is illustrated in Fig.~\ref{fig:pattern_group}. Intuitively, the learning of a cyclic pattern can be considered as learning to extract regions in the frame sequence such that regions containing the same object (the same phase in the cycle) are grouped together. An example is shown in Fig.~\ref{fig:pattern_group}, where the regions containing the trees (red/blue dashed rectangles) are automatically learnt such that regions containing the same type of tree (the same phase of the cycle) belong to the same group. The same process applies to the regions containing the buildings.

\begin{figure}
    \centering
    \includegraphics[width = 0.8\columnwidth]{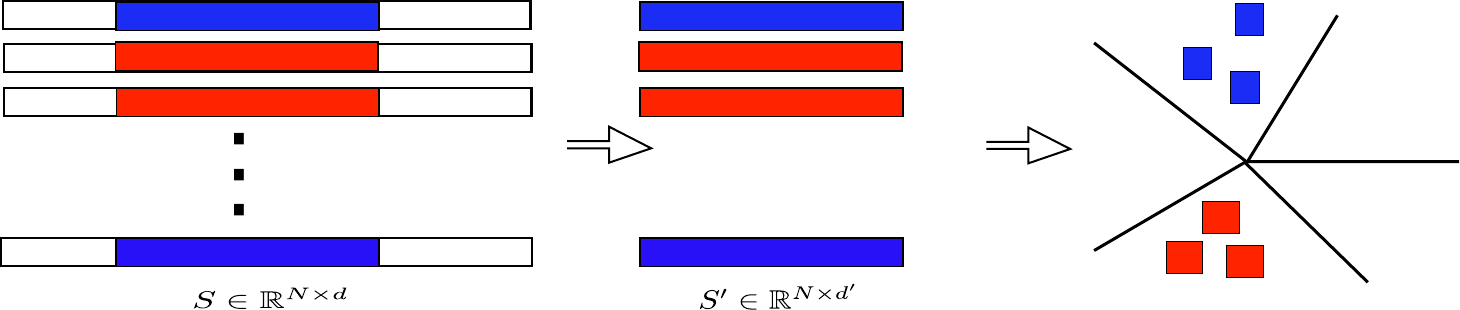}
    \caption{Illustration of pattern learning by selecting a subset of $d'$ columns in $S$ to form a new matrix $S'$ such that the sum of the within-group distances are minimized.} \vspace{-0.5cm}
    \label{fig:matrix_selection}
\end{figure}
Let us denote the set of training data $\{s_1, \dots, s_N\}$ -- each scene $s_i$ is represented by a d-dimensional vector -- by the matrix $S \in \mathbb{R}^{N\times d}$ with $N$ rows and $d$ columns, where each row of $S$ contains one training scene of the dataset. Assume that the phase labels for each cyclic pattern are assigned as described in Sec.~\ref{sec:database_encoding}. Generally speaking, for each cyclic pattern, the learning of the regions as described in Fig.~\ref{fig:pattern_group} can be considered as selecting a subset of $d'$ columns in $S$ ($d' \le d$) to form a new matrix $S' \in \mathbb{R}^{N\times d'}$. This is illustrated in Fig.~\ref{fig:matrix_selection}, in which the colored part of the matrix $S$ represent the columns that need to be extracted to form the matrix $S'$ such that rows assigned with the same phase label (having the same color) are grouped together. In this work, the value of $d'$ is a hyper parameter that is set during the training of the model. The choice of $d'$ affects the storage required for encoding (larger values of $d'$ require more storage). Note that for $k$ cyclic patterns, $k$ different $S'$ matrices are learnt. 

The task of selecting the matrix $S'$ can be done by learning a weight vector $w$, where $w$ has $d$ elements. The value $w_i$ ($1 \le i \le d$) represents the contribution of the $i$-th column into determining the pattern. Specifically, for each column of $S$, the larger the weight value $w_i$, the more significant it is in representing the templates of the cyclic pattern.  Therefore, after the vector $w$ is learnt, we select $d'$ columns with the largest weights. As rows containing the same phase assignment must be in the same group, we would like to select $S'$ such that the sum of the within-group distances of the training data is minimized. Mathematically speaking, given the cycle length $\tau$, and the label assignments $l_1,\dots, l_N$, the weight vector $w$ is learnt by solving the following optimization problem
\begin{equation}
    \label{eq:w_learning}
    \begin{aligned}
    & \min_w &&\sum_{i = 1}^{\tau} \sum_{j = 1}^{d} \sum_{p,q \in \hat{S}_i} w_j (p_j - q_j)^2\\
    & \text{subject to} && w^2_1 + \dots + w^2_d \le 1 \\
    & && |w_1| + \dots + |w_d| \le \gamma, \;\;\;
    w_j \ge 0 \;\; \forall j
    \end{aligned}
\end{equation}
where the set $\hat{S_i}$ contains the data points having the same label $i$, i.e., $\hat{S}_i = \{s_j \in S | l_j = i \}$, and $\gamma$ is a tuning parameter, which is chosen in the range of (0.1, 100) in our experiments.
Intuitively, by minimizing~\eqref{eq:w_learning}, one is looking for a vector $w$ that minimizes the sum of the within-group squared distances, which is illustrated in Fig.~\ref{fig:matrix_selection}.

Our pattern learning technique is inspired by the class of feature selection methods that are commonly used for K-Means clustering~\cite{witten2010framework}. In our work, however, instead of applying K-Means, we assume that the scenes having the same phase labels are already clutered into the same group. Our task is to learn the weight vector $w$ that minimizes the within cluster sum of squares.

Similar to~\cite{witten2010framework}, in order to solve~\eqref{eq:w_learning}, we can maximize the between-cluster sum of squares (BCSS), which can be written as
\begin{equation}
    \label{eq:w_learning_max}
    \begin{aligned}
    & \max_w && \sum_{j = 1}^{d} w_j \left( \frac{1}{N}\sum_{p,q \in S} (p_j - q_j)^2 -  \sum_{i=1}^{\tau}\frac{1}{|\hat{S}_i|}\sum_{p,q \in \hat{S}_i}  (p_j - q_j)^2 \right)\\
    & \text{s.t.} && w^2_1 + \dots + w^2_d \le 1 \\
    & && |w_1| + \dots + |w_d| \le \gamma, \;\;\;
    w_j \ge 0 \;\; \forall j
    \end{aligned}
\end{equation}
The solution to the problem~\eqref{eq:w_learning_max} can be computed using soft-thresholding, as described in~\cite[Proposition 1]{witten2010framework}.
\subsection{Encoding storage analysis}
\label{sec:encoding_storage}
In this section, we analyze the main storage required for our encoder. As described in Sec.~\ref{sec:database_encoding} and Sec.~\ref{sec:pattern_learning}, the following storage is required by our encoding algorithm:
\begin{itemize}
    \item \textbf{Phase encoders:} For a phase encoder with cycle period of $\tau$, as we train our phase encoders using linear SVM as in~\cite{yu2018rhythmic}, we need to store $\tau$ set of parameters, where each set contains $d'+1$ real numbers ($d'$ numbers representing the hyperplane and $1$ value for the bias). Here, $d'$ is the length of the vector representing the templates of the cyclic patterns. The value of $d'$ is specified during template learning, as described in Sec.~\ref{sec:pattern_learning}. Note that by chosing $d' \ll d$, the storage required by phase encoders in our algorithm is much less than the amount of storage required by~\cite{yu2018rhythmic}, while the precision is maintained. In our experiments, the value of $d$ is chosen to be $d' = \rho d$, where $\rho$ ranges from $40\%$ to $60\%$. 
    \item \textbf{Feature selection for pattern learning:} As described in Sec.~\ref{sec:pattern_learning}, from the $d$-dimensional feature vectors stored in the matrix $S \in \mathbb{R}^{N\times d}$, for each cyclic pattern, we need to select $d'$ column of $S$ to train the phase encoder. After learning the weight vector $w$, the indexes of the selected columns can be encoded using a binary vector containing $d$ bits, where $d'$ bits corresponding to $d'$ are set to $1$, while the remaining bits are all zeros.  
\end{itemize}
In summary, with $k$ cyclic patterns with cycle periods of $\tau_1, \dots, \tau_k$, the total storage required (in bytes) can be calculated as
\begin{equation}
    8(d'+1)(\tau_1+\dots+ \tau_k) + k\frac{d}{8} \;\;\text{(bytes)}
\end{equation}
Here we assume that each real number is stored using 64 bits (8 bytes). The same computation can be applied for systems that use 32 bits (4 bytes).
\subsection{Sequence chunking for large scale datasets}
\label{sec:sequence_chunking}
In the previous sections, we have described our method for encoding a dataset using repeated cyclic patterns. In practice, however, the traversal of a robot or a vehicle may comprise different types of environments that contain totally different cyclic patterns. For instance, when a car enters a tunnel, the templates containing trees and buildings are no longer available. Instead, the localization may depend on the cycles of the landmarks on the wall of the tunnel or on the road.  In such scenarios, the use of only one phase encoder that encode all the templates belonging to different patterns, e.g., trees and wall landmarks, may degrade the overall performance as different classes are assigned with the same label during the training of the encoders. 
\begin{figure}[ht]
    \centering
    \includegraphics[width=0.35\textwidth]{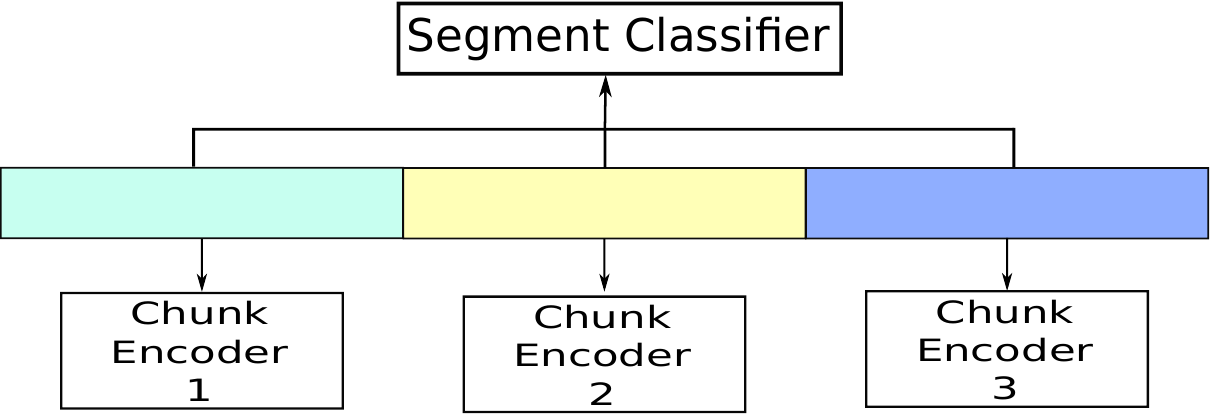}
    \caption{Illustration of the chunking process for a dataset with 3 different chunks. Each chunk is encoded separately using the method described in Sec.~\ref{sec:database_encoding} and Sec.~\ref{sec:pattern_learning}}
    \label{fig:chunking}
\end{figure}
To address this issue, we propose a dataset partition strategy which enables our algorithm to work with large scale datasets. Fig.~\ref{fig:chunking} illustrates an example of our chunking process where a dataset is partitioned into three data chunks. By segmenting the whole sequence into different parts, different environments containing different types of cyclic patterns can be encoded separately using the method described in the previous sections. In Fig.~\ref{fig:chunking}, three different chunks are shown in three different colors, and each individual chunk is encoded using a separate chunk encoder. 

Since the whole dataset is split into multiple data chunks that are encoded separately, in order to identify which chunk encoder to use, we introduce the segment classifier (or chunk classifier, which is also shown in Fig.~\ref{fig:chunking}). This chunk classifier learns from all the input data to identify which chunk a scene belongs to. Similar to the phase encoders, we also use linear SVM to train the chunk classifier, where the label for each frame is the index of the chunk to which it belongs. The same pattern learning approach as described in Sec.~\ref{sec:pattern_learning} can also be applied to reduce the dimensionality and the required storage for the chunk classifier.

To localize a new frame, it is first passed into the chunk classifier to identify the data chunk containing the frame. After the chunk index is known, that particular chunk encoder can be used for precise localization. 

Note that with the introduction of data chunking, besides the storage required for the chunk encoders, which can be computed as described in Sec.~\ref{sec:encoding_storage}, we also need to store the parameters for the chunk classifier. The additional storage for the chunk classifier is
\begin{equation}
    8C(\tilde{d} + 1 ) + \frac{d}{8} \;\; \text{(bytes)},
\end{equation}
where $C$ is the number of chunks that the dataset is partitioned into and $\tilde{d}$ is the length of the feature vectors extracted from the original data to train the chunk encoder. Note that since the number of chunks do not scale linearly with the number of training scenes, this chunking scheme still maintain the sub-linear growth of the encoding storage.
\section{Experiments and Results}
\label{sec:experiments}
In this section, we describe our experimental setup and the results of testing our algorithm on different real-world datasets. By running our encoding algorithm on different training sizes, we show that the algorithm can achieve sub-linear storage growth as the size of the training dataset increases.  We also compare our algorithm with the work of~\cite{yu2018rhythmic} to demonstrate that our algorithm is able to provide better performance in terms of accuracy while requiring less storage. Additionally, we evaluate the performance of our algorithm and its new chunking capability on very large datasets.

As the main focus of our work is to develop an encoding algorithm with sub-linear growth, we compare our work with~\cite{yu2018rhythmic}, as this is the only work in the literature that can achieve this storage growth requirement. Although other encoding techniques such as~\cite{jegou2011product,ge2013optimized} can also be used for visual place recognition, the storage required by those methods must scale at least linearly with respect to the training storage.

Our algorithm is implemented in Python and tested on an Ubuntu machine with 32GB of RAM at 4.2GHz. 
\subsection{Dataset and Pre-processing}
Two commonly used datasets are selected to evaluate the performance of our algorithm:
\subsubsection{Aerial Brisbane Dataset}This dataset contains images captured from NearMaps\footnote{https://www.nearmap.com.au/} covering the Brisbane region in Queensland, Australia. From the original snapshot with the size of 7562 $\times$ 6562, where each pixel represents an actual geographic area of 4.777 $\times$ 4.777 square meters, it is segmented in to $224 \times 224$ frames with 112-pixel strides, resulting in 3075 frames per dataset. Different snapshots were collected to evaluate the performance of the algorithm under varying visual environments. In our experiments, we combine 4 different snapshots for training and 4 other different snapshots for testing, resulting in a training dataset and a testing dataset, each containing 12300 frames.
\subsubsection{Nordland Train Dataset} This dataset\footnote{https://nrkbeta.no/2013/01/15/nordlandsbanen-minute-by-minute-season-by-season/} is collected from a front-facing camera installed at the front of a train running for 10 hours. Four video sequences are collected through four seasons of the year: fall, summer, spring and winter. This is a large scale dataset that has been used extensively throughout multiple works on visual place recognition, which provides challenging visual changes of the frames. Each video sequence contains more than 890,000 frames.  To evaluate the performance of the algorithms with medium and large-scale datasets, we sampled 8,900 frames and 89,000 frames to create two datasets: Nordland-8K and Nordland-80K respectively.

Our encoding algorithm can be used on top of different state-of-the-art feature extractors for visual place recognition. In this work, we use LoST~\cite{garg2018lost} to generate features for the training and testing datasets. All the images are fed into the pre-trained network provided by the authors\footnote{https://github.com/oravus/lostX} to obtain the LostX features that represent the images, where each image is represented by a vector of 8192 dimensions. The data is then normalized, which are then used to train and test our model. For experiments where testing data is identical to the training data, we also conduct PCA to reduce the data dimensions to 5000 in order to reduce training and testing time.

The storage required of training datasets reported in our experiments are computed based on the number of frames ($N$) and the length of the the feature vector ($d$) that represents an image. The encoding storage is computed as described in Sec.~\ref{sec:encoding_storage}. We assume that each real number in the descriptor vector is stored using 64 bits (8 bytes). Values of storage size for all the experiments are reported in megabytes (MB). We used the latest version of LIBLINEAR~\cite{fan2008liblinear} to implement our phase encoders, which allows the training process to be conducted on multiple cores.

\subsection{Sub-linear scale analysis}
The aim of this experiment is to evaluate whether our encoding algorithm is able to achieve sub-linear growth of the encoding storage as the number of scenes in the training dataset increases, \emph{while maintaining the same prediction precision}. As with the original study~\cite{yu2018rhythmic}, we conduct two sets of experiments: a diagnostic set of experiments when the testing and training datasets are identical, and experiments where the test datasets exhibit the challenging appearance changes caused by varying time of day and weather conditions.

\subsubsection{Experiments with identical testing and training datasets}

To simulate the increase in size of training data, we repeatedly sample $N$ frames from the Nordland Fall and Brisbane dataset, with N increases from 100 to around 10K frames. For each value of $N$, we run our encoding algorithm and record the smallest encoding storage required to maintain the same precision. The parameters are chosen such that the methods achieve the prediction precision of 90\% and 80\%.  

\begin{figure}[ht]
    \centering
    \includegraphics[width=0.49\columnwidth]{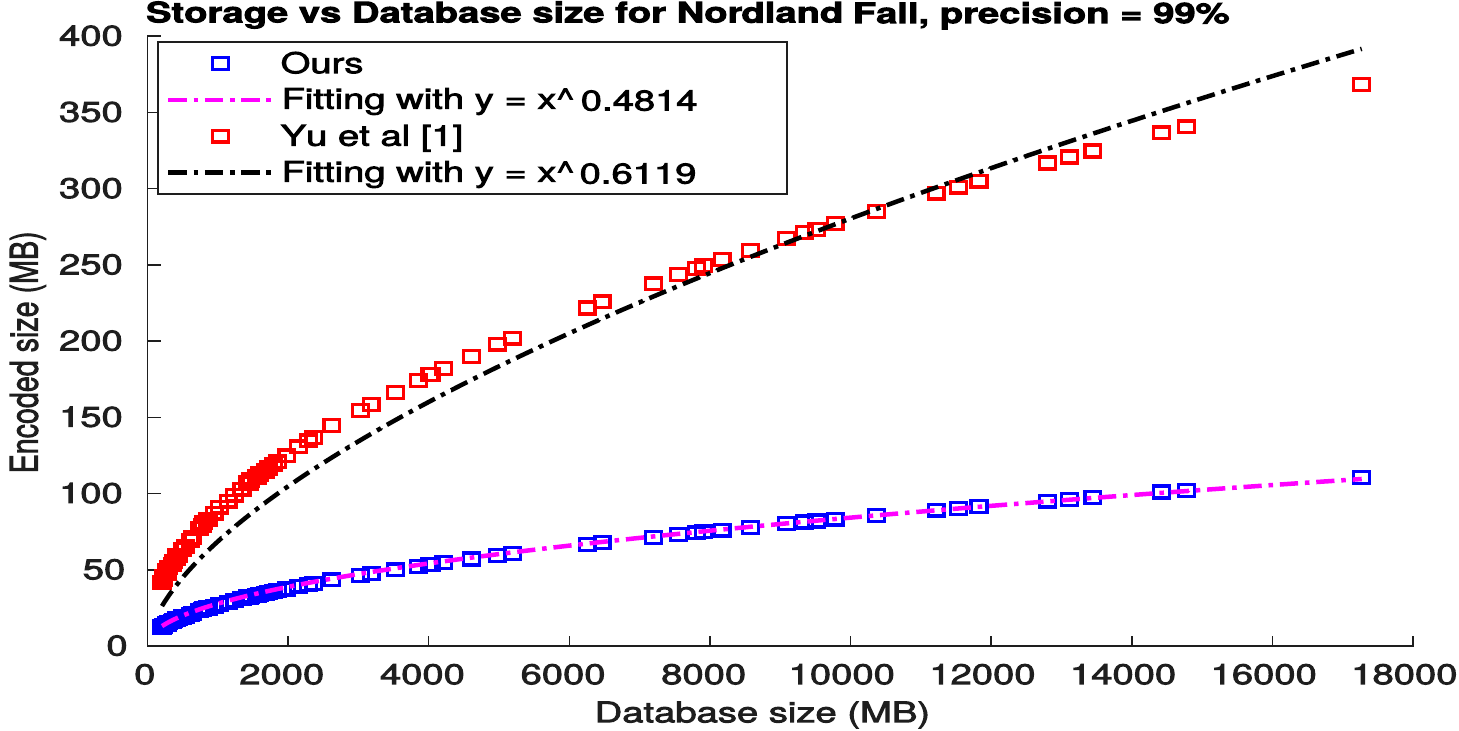} 
    \includegraphics[width=0.49\columnwidth]{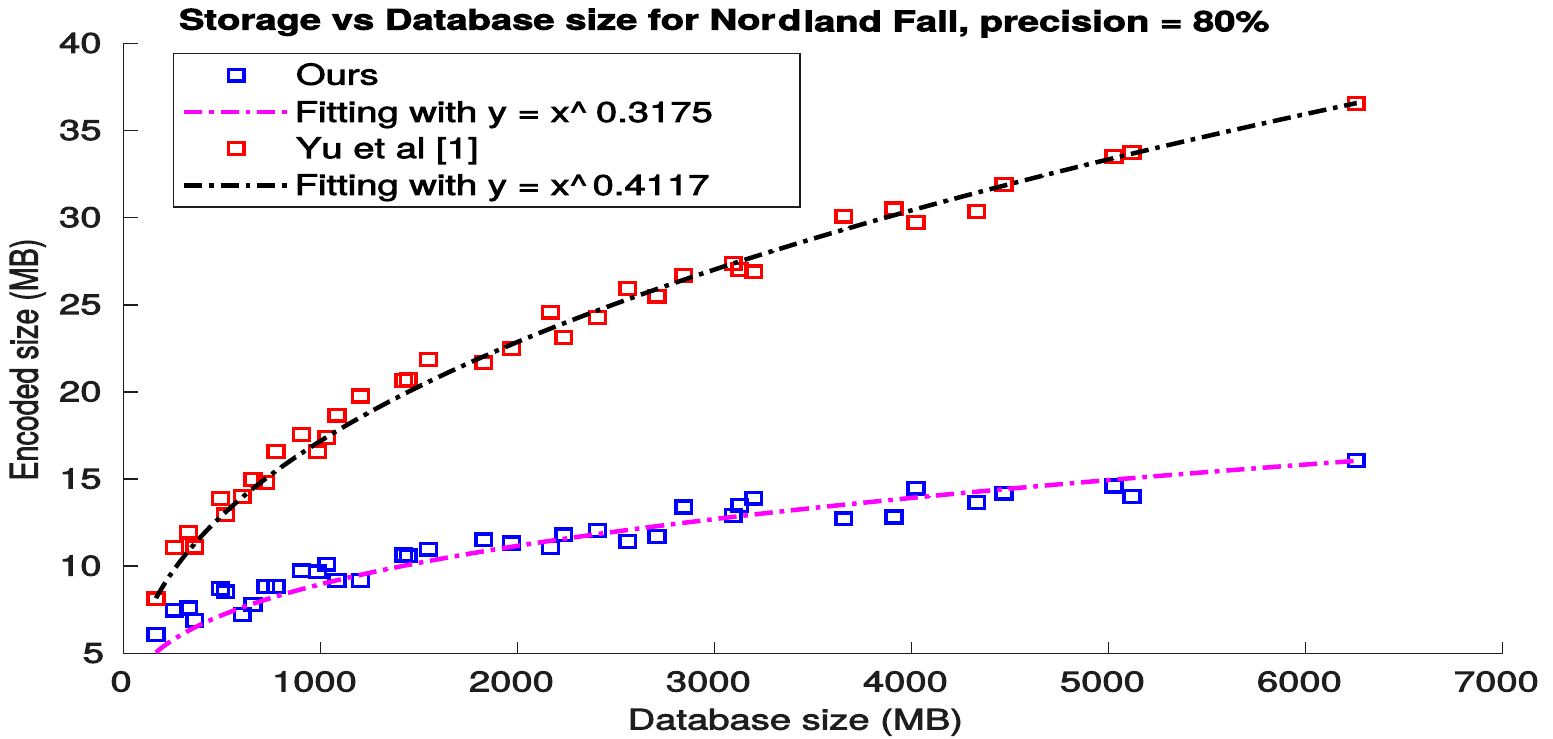} 
    \includegraphics[width=0.49\columnwidth]{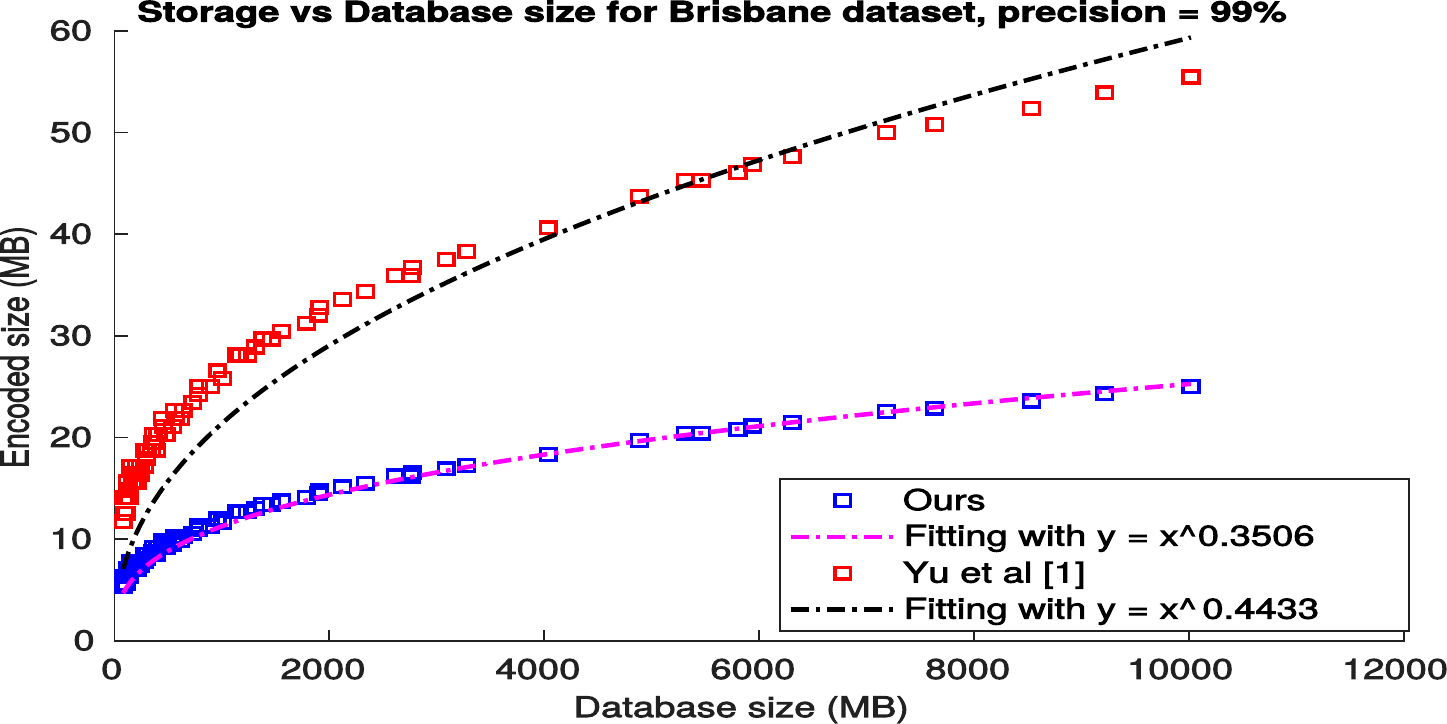}
    \includegraphics[width=0.49\columnwidth]{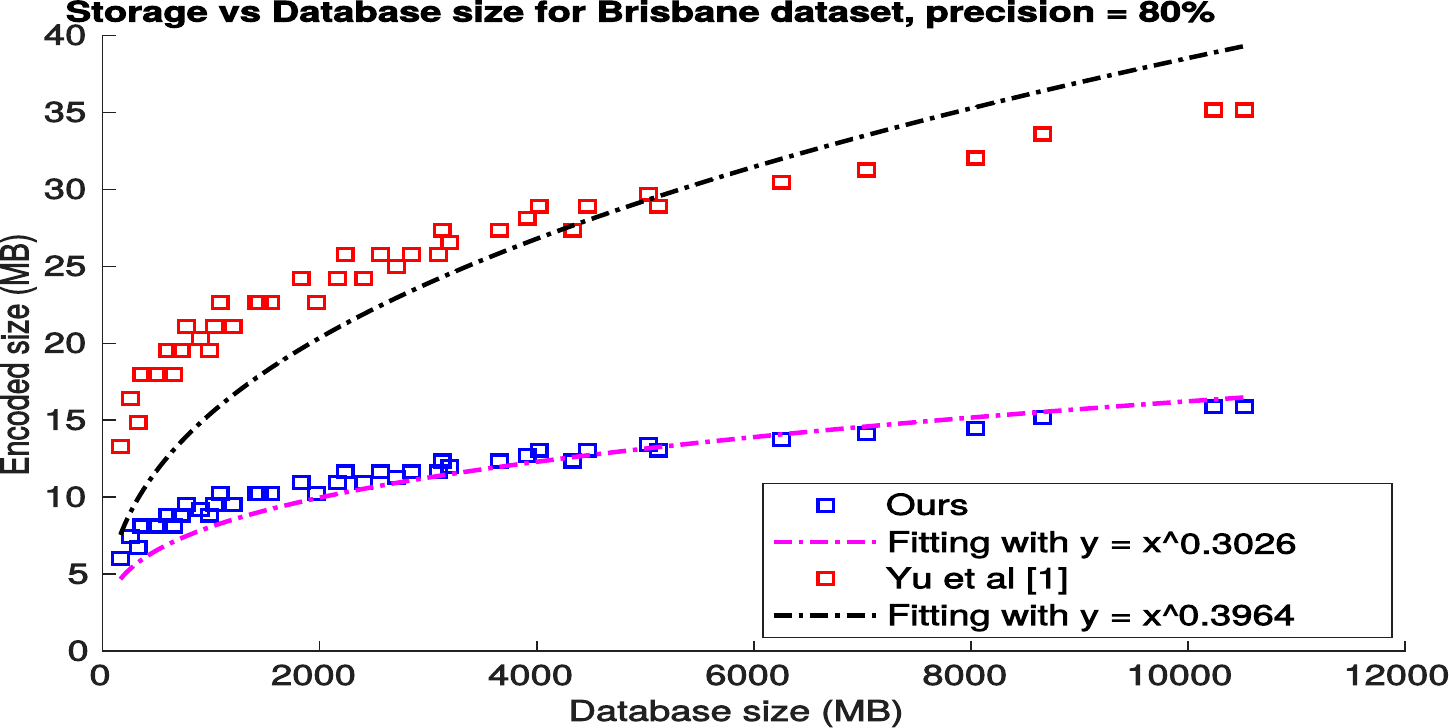}
    \caption{Scale of storage required as the size of the training dataset increases for Nordland Fall dataset (top) and Brisbane dataset (bottom) when the testing dataset is identical to the training dataset to maintain the same precision of $99\%$ (left) and $80\%$ (right). }
    \label{fig:scale_identical}
\end{figure}
Fig.~\ref{fig:scale_identical} plots the storage required versus training size, where the testing data is identical to the training data for Nordland and Brisbane dataset, respectively. As can be seen from this figure, the amount of storage required by our algorithm is significantly less than that of~\cite{yu2018rhythmic}, while the same prediction precision is maintained. This is achieved by our effective pattern learning algorithm, as redundant information about the cyclic patterns are removed, and the remaining information is still sufficient to train the phase encoders such that the templates of the cycle are precisely classified.

To investigate the characteristics of the sub-linearity, we perform a curve fitting (using MATLAB's curve fitting toolbox) to the set of data points on the graphs with the sub-linear function $y = x^a$ (with $a<1$), where $y$ represents the storage size and $x$ represents the database size. Table.~\ref{table:a_values_same} summarizes the value of $a$ with different experiment settings. Note that the $a$ values of our method is smaller than that of~\cite{yu2018rhythmic}, as our method requires less storage. Also, as the required precision reduces, the value of $a$ is also reduced. This shows the trade-off between the storage size and the localization accuracy. Our method enables the user to easily adjust the storage required based on the specific applications, an improvement over the rigid co-prime requirements of the original approach~\cite{yu2018rhythmic}.
\begin{table}[]
    \centering
    \caption{Fitting values of $a$ for the function $y = x^a$ (with $a<1$). }
    \begin{tabular}{|c|c|c|c|c|c|}
    \hline
    \multicolumn{1}{|c|}{\multirow{2}{*}{Train}} & \multirow{2}{*}{Test} & \multicolumn{2}{c|}{Precision} & \multicolumn{2}{c|}{Values of a} \\ \cline{3-6} 
    \multicolumn{1}{|c|}{} &  & Ours & {}\cite{yu2018rhythmic}{} & Ours & {}\cite{yu2018rhythmic}{} \\ \hline
    Nordland Fall & Nordland Fall & 99\% & \multicolumn{1}{c|}{99\%} & \textbf{0.4814} & 0.6119 \\ \hline
    Nordland Fall & Nordland Fall & 80\% & 80\% & \textbf{0.3175} & 0.4117 \\ \hline
    Brisbane 1 & Brisbane 1 & 99\% & 99\% & \textbf{0.3505} & 0.4433 \\ \hline
    Brisbane 1 & Brisbane 1 & 80\% & 80\% & \textbf{0.3026} & 0.3946 \\ \hline
    Nordland Fall & Nordland Summer & \textbf{50\%} & \multicolumn{1}{c|}{30\%} & \textbf{0.6481} & 0.7133 \\ \hline
    Brisbane 1 & Brisbane 2 & \textbf{60\%} & 50\% & \textbf{0.6479} & 0.7173 \\ \hline
    \end{tabular}
    \label{table:a_values_same}
\end{table}

\subsubsection{Experiments with visually changing testing datasets}
We also evaluate the sub-linearity growth of the required storage to maintain the \emph{same localization precision} under different testing conditions. The same experiment as in the previous section is repeated, but the testing dataset is visually different to the training dataset due to weather and time-of-day changes.
\begin{figure}
    \centering
    \includegraphics[width=0.49\columnwidth]{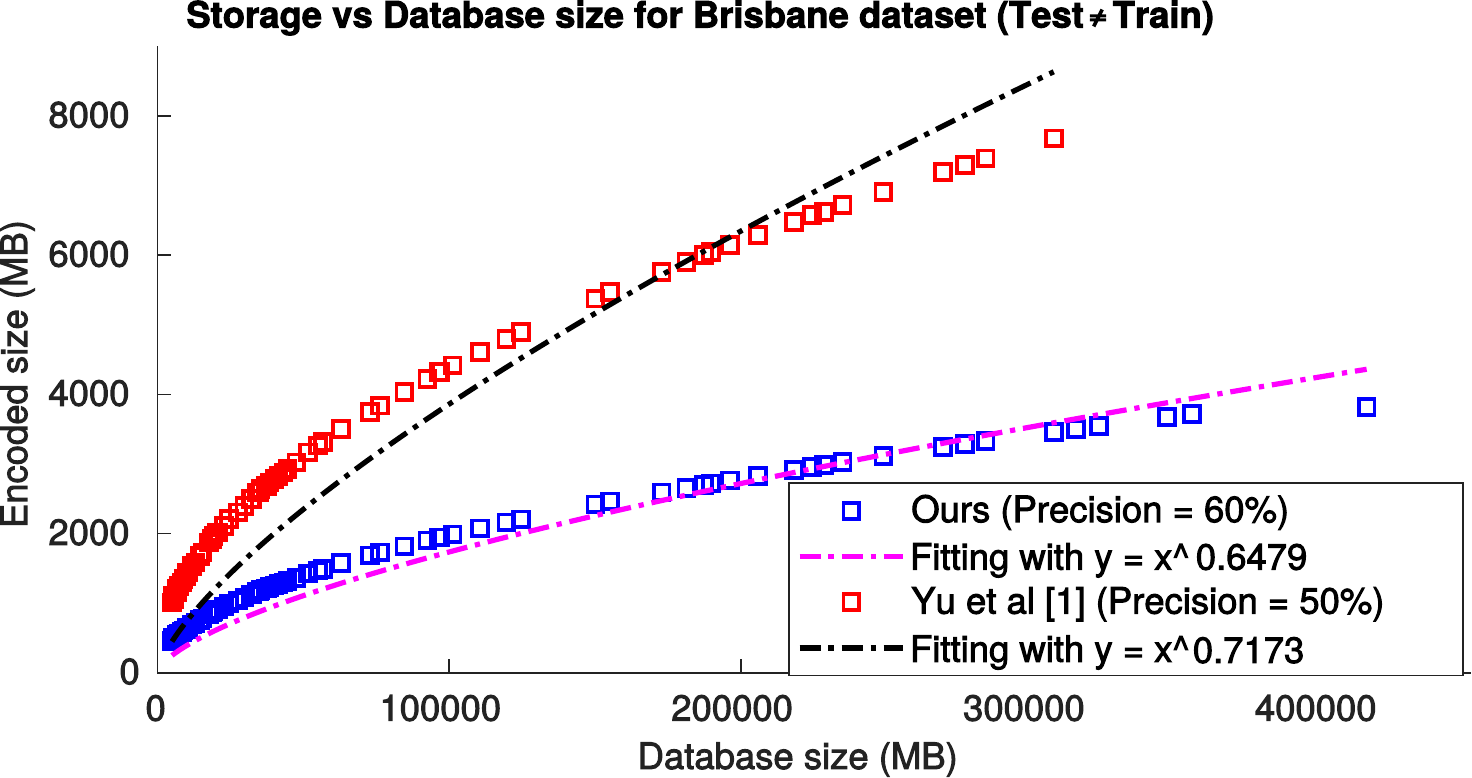}
    \includegraphics[width=0.49\columnwidth]{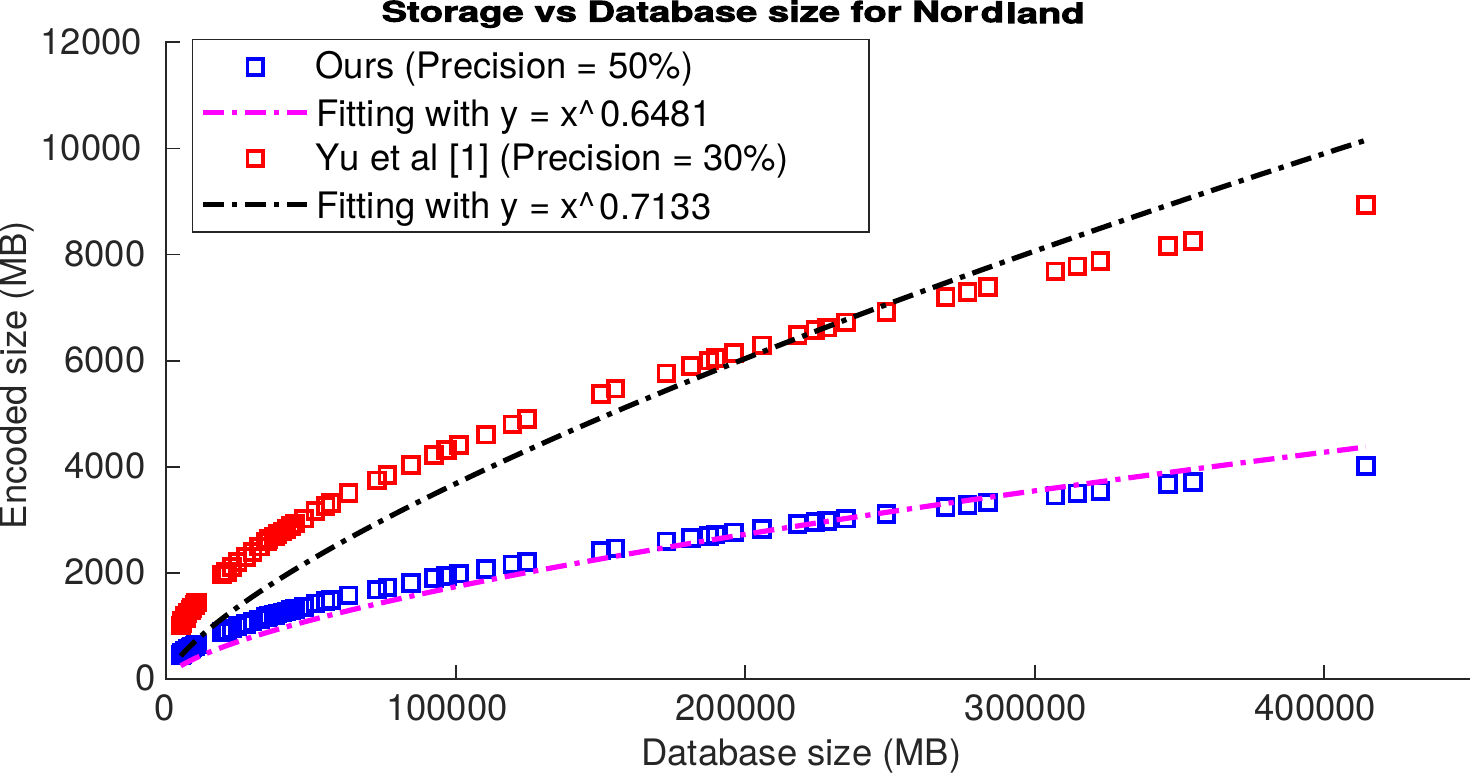}
    \caption{Scale of storage required as the size of training dataset increases for when the testing dataset is different from the training dataset. Left: Brisbane with precision of $60\%$ (our method) and $50\%$ (method of~\cite{yu2018rhythmic}). Right: Nordland with precision of $50\%$ (our method) and $30\%$ (method of~\cite{yu2018rhythmic}). Our method achieves higher precision, while the amount of storage required is much less than~\cite{yu2018rhythmic}. The precision is measure with tolerance of 5 frames.} \vspace{-0.5cm}
    \label{fig:scale_diff}
\end{figure}

Fig.~\ref{fig:scale_diff} shows experimental results for the Nordland dataset and Brisbane dataset. For Nordland, the models were trained on Nordland Fall dataset and tested on the Nordland Summer dataset with the same sequence of frames. With this challenging dataset, our method is able to achieve a precision of $50\%$, while the method of~\cite{yu2018rhythmic} can only achieve a precision of $30\%$ and requires a much higher amount of encoding storage. For Brisbane, our method achieves the precision of $60\%$, superior to the precision of $50\%$ achieved by the original method~\cite{yu2018rhythmic}. Note that our new method is also able to obtain a higher precision with significantly less storage compared to~\cite{yu2018rhythmic}.

\subsection{Training and Testing time}
With the new encoding scheme, we can achieve not only lower storage, but also faster training and querying time. In Table~\ref{table:training_testing_time}, we show the average training and testing time for different values of $N$ (database size). Note that the parameters for these experiments are chosen so that the two methods provide the same localization precision. Both methods are implemented with the same SVM library (LIBLINEAR). As can be seen from Table~\ref{table:training_testing_time}, our method achieves faster training and testing time compared to~\cite{yu2018rhythmic}. This is the result of our pattern learning approach, which allows the phase encoder to be trained with much less data compared to~\cite{yu2018rhythmic}. Also, due to the new phase label assignment scheme, the query time is significantly faster as our method do not require the set intersection computation as proposed by~\cite{yu2018rhythmic}.
\begin{table}[ht]
    \centering
     \caption{Comparison of training and testing time for different values of dataset size between our method and~\cite{yu2018rhythmic}}
    \begin{tabular}{|c|c|c|c|c|}
    \hline
    \multirow{2}{*}{N} & \multicolumn{2}{l|}{Training Time (s)} & \multicolumn{2}{l|}{Query Time (s)} \\ \cline{2-5} 
     & Ours & Yu et al.~\cite{yu2018rhythmic} & Ours & Yu et al.~\cite{yu2018rhythmic} \\ \hline
    1000 & \textbf{9.83} & 22.14 & \textbf{2.79} & 6.11 \\ \hline
    2107 & \textbf{78.71} & 106.54 & \textbf{8.40} & 13.28 \\ \hline
    4288 & \textbf{187.74} & 406.76 & \textbf{17.01}  & 26.49 \\ \hline
    5183 & \textbf{282.91} & 485.51 & \textbf{21.36} & 32.82 \\ \hline
    6000 & \textbf{346.28} & 600.95 & \textbf{23.12} & 38.36 \\ \hline
    \end{tabular} \vspace{-0.5cm}
    \label{table:training_testing_time}
\end{table}

\subsection{Performance analysis with varying number of patterns}
As the main mechanism behind our encoding algorithm relies on the use of cyclic patterns, we investigate in this experiment how the choice of the number of cyclic patterns affects the performance of our algorithm. The method of~\cite{yu2018rhythmic} is also put into comparison. 
\subsubsection{Experiments with identical testing and training datasets}
\label{sec:exp_pattern_same}
\begin{figure}
    \centering
    \includegraphics[width=0.49\columnwidth]{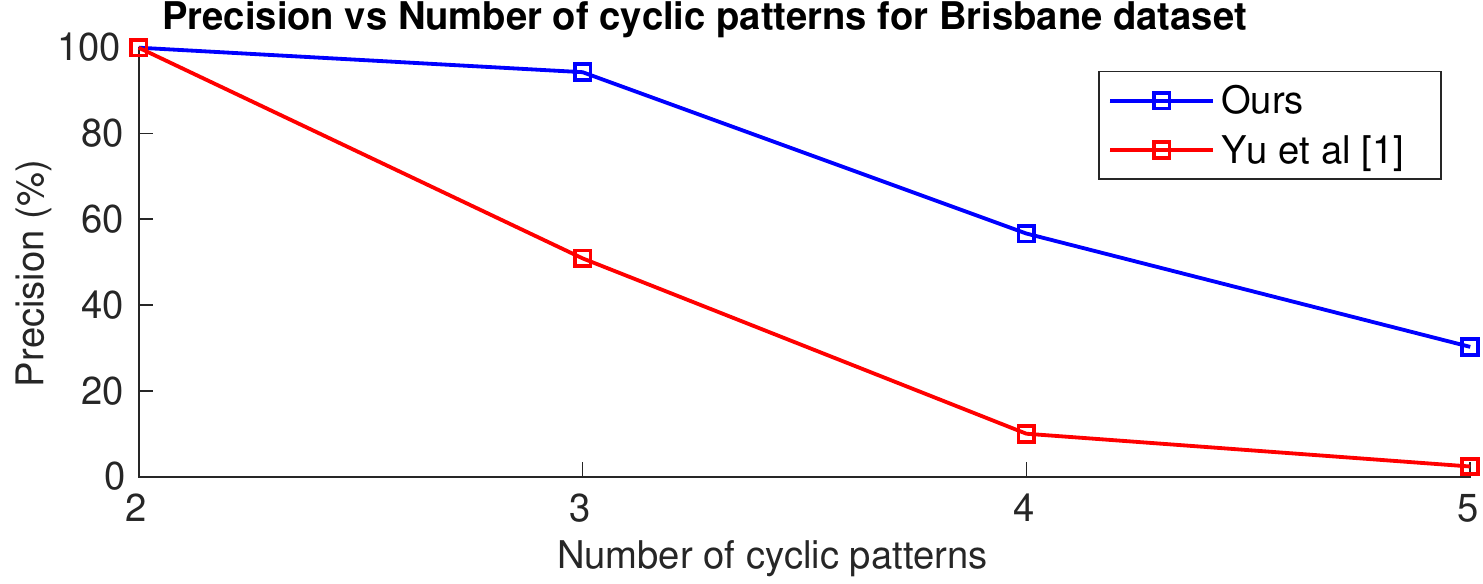}
    \includegraphics[width=0.49\columnwidth]{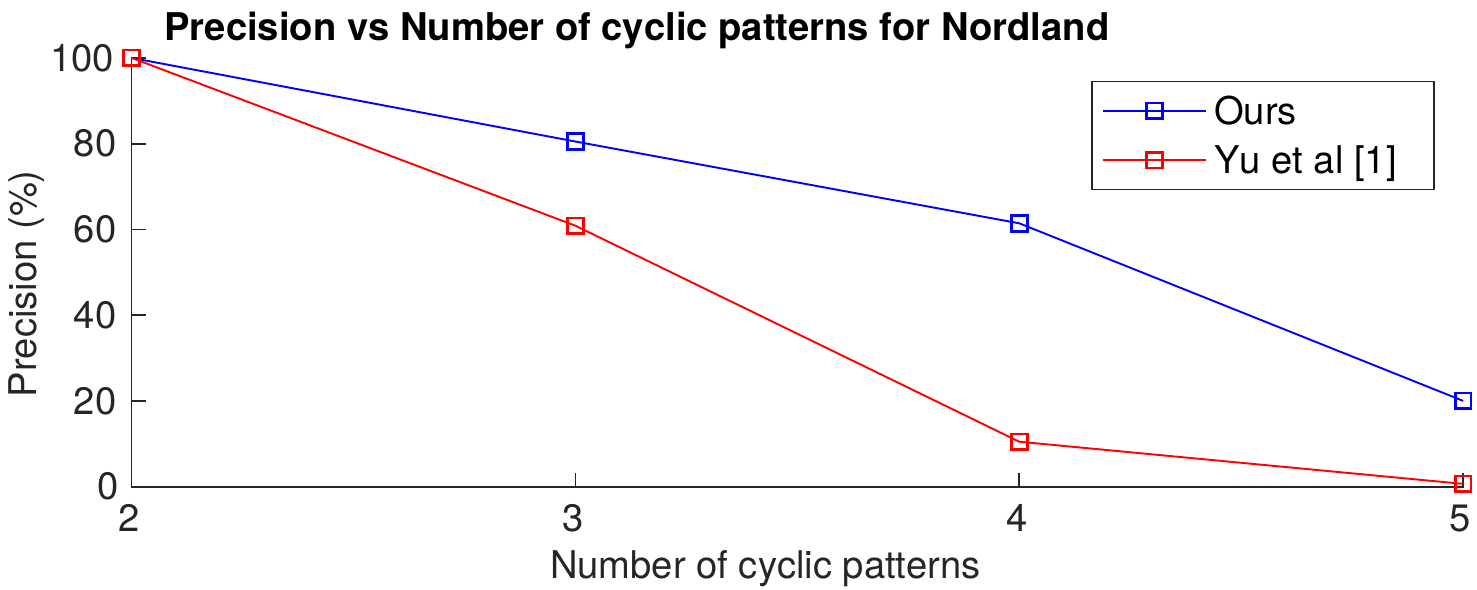}    
    \caption{Localization precision under different number of cyclic patterns for Brisbane and Nordland Summer dataset when testing dataset is identical to training dataset. Our method is able to maintain good precision, while the precision achieved by~\cite{yu2018rhythmic} drops substantially as the number of pattern increases. } \vspace{-0.5cm}
    \label{fig:pattern_Nordland_brisbane}
\end{figure}

We used 8000 frames from the Nordland Summer dataset and run our encoding algorithm where the number of cyclic patterns changes from $k=2$ to $k=5$ and record the localization precision. For each value of $k$, the cycle lengths are chosen to be approximately $\sqrt[k]N$ and the same values of cycle lengths are used for both our method and the method of~\cite{yu2018rhythmic}. The results are shown in Fig.~\ref{fig:pattern_Nordland_brisbane}. When the number of patterns increases, our method can achieve relatively good performance compared to the method of~\cite{yu2018rhythmic}. Note that the localization precision of~\cite{yu2018rhythmic} drops substantially as the number of patterns increases.

\subsubsection{Experiments with visually changing datasets}
\begin{figure}[ht]
    \centering
    \includegraphics[width=0.49\columnwidth]{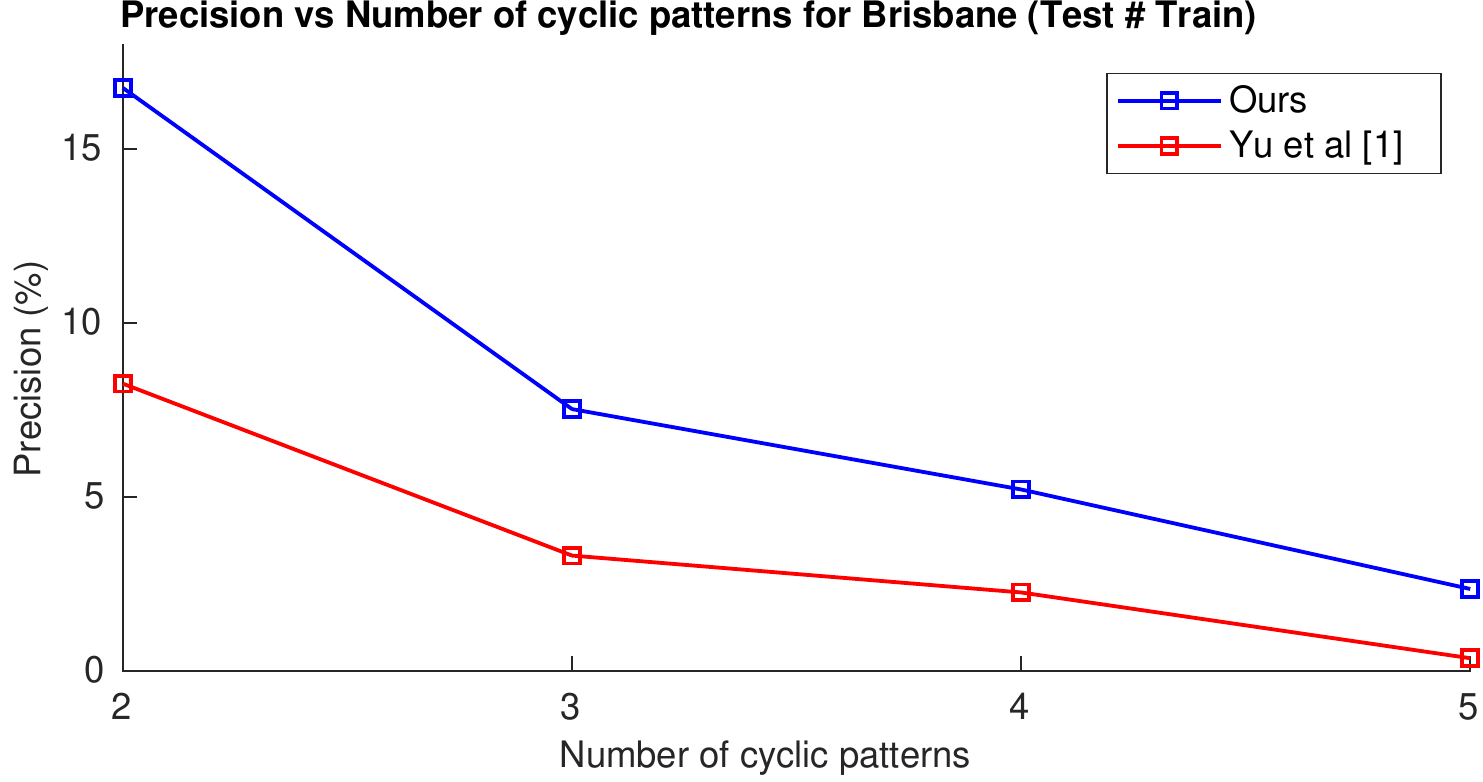}
    \includegraphics[width=0.49\columnwidth]{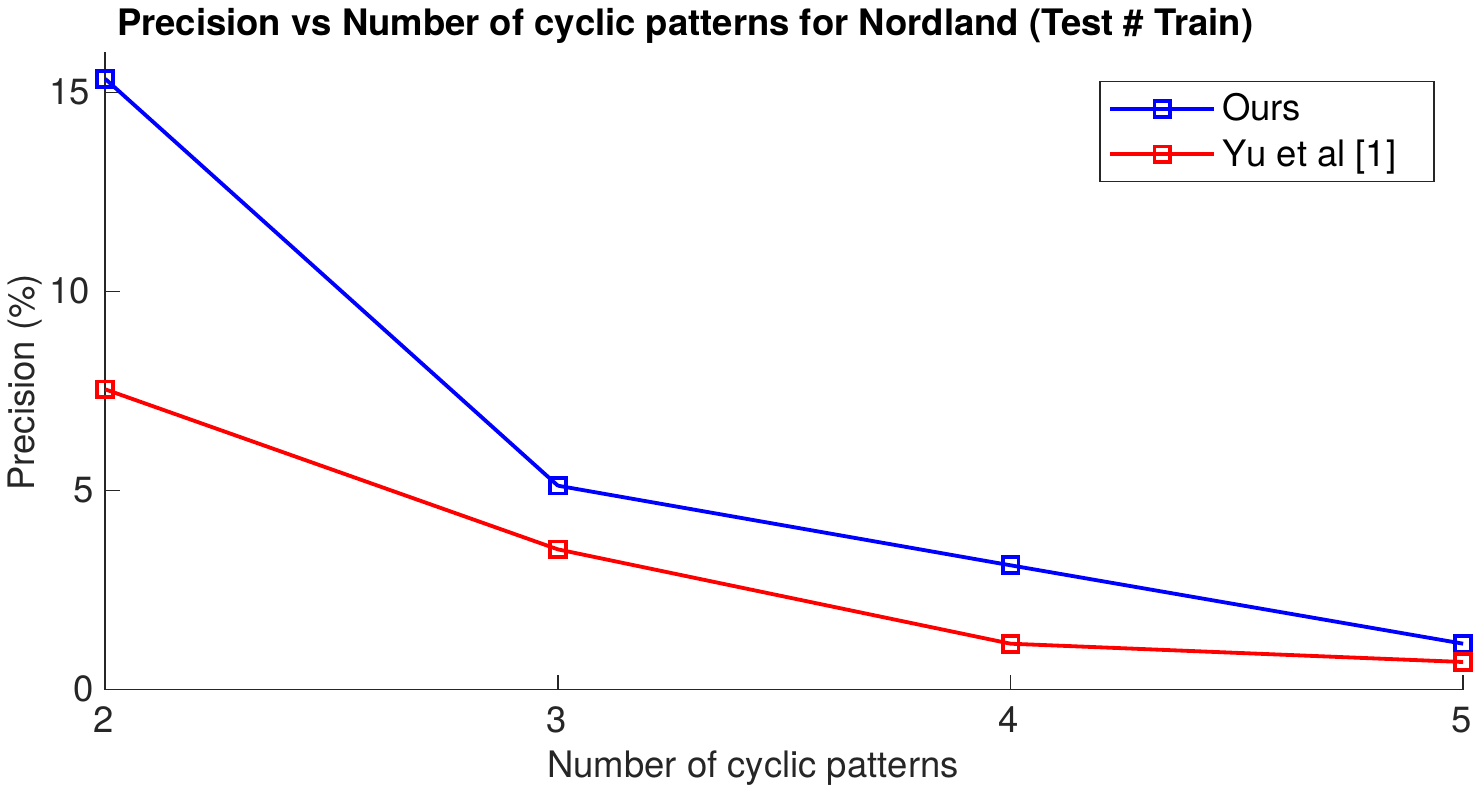}    
    \caption{Localization precision under different number of cyclic patterns for Brisbane and Nordland dataset when testing dataset is different from training dataset. Our method is able to maintain good precision, while the precision achieved by~\cite{yu2018rhythmic} drops substantially as the number of pattern increases. }
    \vspace{-0.5cm}
    \label{fig:pattern_Nordland_brisbane2}
\end{figure}

The same experiment as in Sec.~\ref{sec:exp_pattern_same} is repeated where we used 1000 frames of Nordland Fall to train our model, and tested on 1000 frames of Nordland Summer dataset (with the same sequence order). For Brisbane, we trained on 1000 frames of Brisbane 1 and tested on 1000 other frames of Brisbane 2. The results are shown inf Fig.~\ref{fig:pattern_Nordland_brisbane2}. Similar to Fig.~\ref{fig:pattern_Nordland_brisbane}, throughout different values of cyclic patterns, the precision is maintained better by our method, while the performance of~\cite{yu2018rhythmic} degrades drastically with the increase of pattern number.

\subsection{Sequence chunking for large scale dataset}
In this section, we benchmark the performance of our sequence chunking strategy described in Sec.~\ref{sec:sequence_chunking} and show that our approach is able to work with large scale datasets that contain up to approximately 90,000 frames. 

\begin{figure}[ht]
    \centering
    \includegraphics[width=0.8\columnwidth]{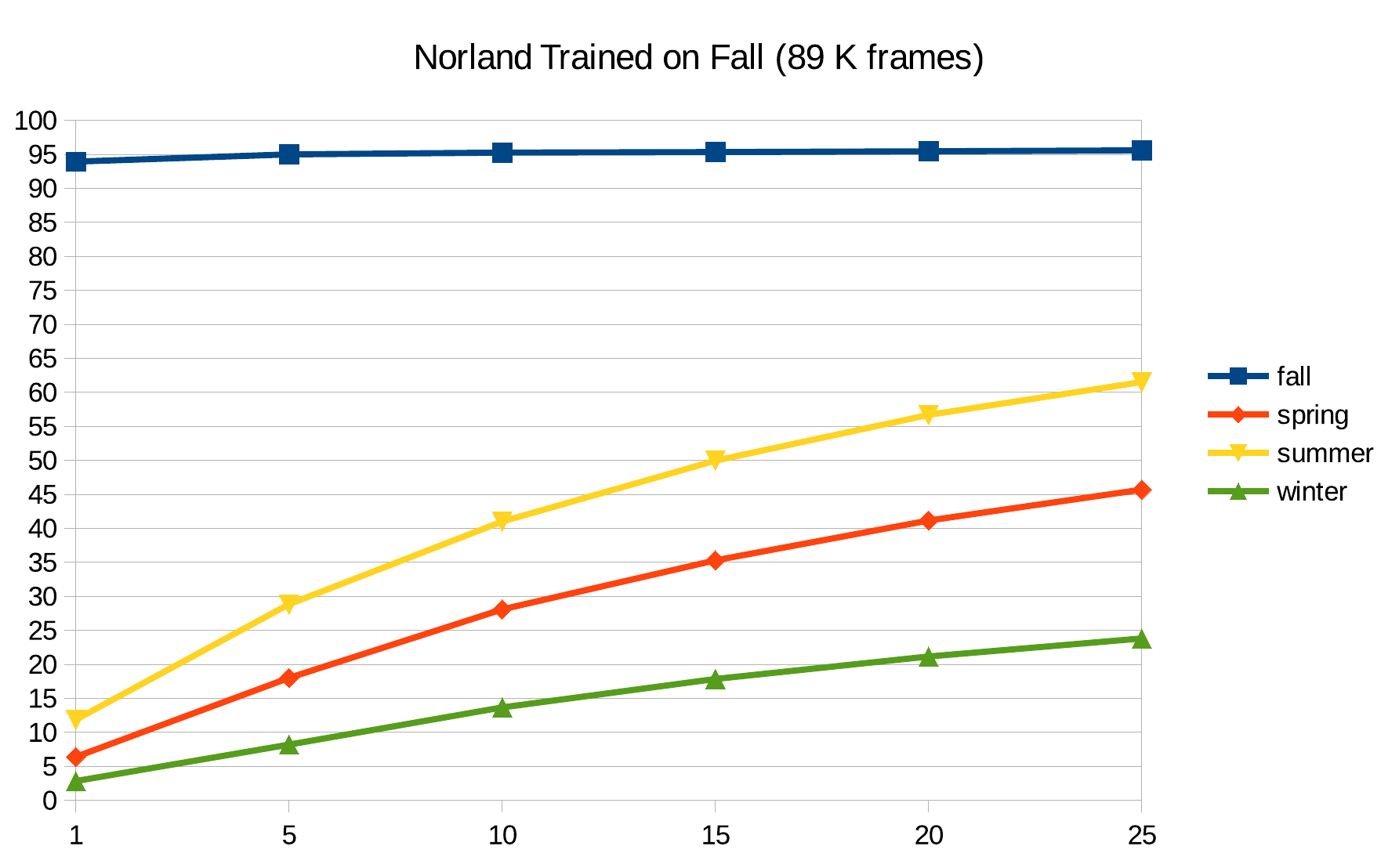}
    \caption{Localization results for Nordland dataset with 89,000 frames. The model is trained on Nordland Fall dataset. } \vspace{-0.5cm}
    \label{fig:Nordland_80k}
\end{figure}

For each season in the Nordland dataset (Fall, Summer, Winter, Spring), we sub-sampled 89,000 frames. The frames from the Fall season are used to train the model. The input frames are partitioned into 10 chunks, each chunk contains 8900 frames. These 10 chunks are trained separately using 10 chunk encoders. The trained model is then tested with frames from the remaining seasons. Fig.~\ref{fig:Nordland_80k} plots the precision when tested on different datasets over increasing frame tollerance. As can be seen, with this large dataset and the chunking approach, we get very high precision (95\%) when we test the model on Nordland Fall dataset (which is identical the the training dataset). When tested with other datasets, we get relatively good performance. When the frame tolerance is allowed to be 25 frames (which is reasonable in practice), our system can achieve the precision of 60\% when tested on Summer (slight condition change). With Spring and Winter, the precision drops, as the scenes undergo larger changes. 

\section{Discussion and Future Work}
\label{sec:conclusion}

We have presented a novel encoding method for visual place recognition, where the storage required scales sub-linearly with respect to the size of the training data. A new pattern learning technique is embedded into our framework, enabling our method to encode only the visual data directly relevant to the encoded periodic patterns, leading to a factor reduction in absolute storage requirements compared to the previous work~\cite{yu2018rhythmic} while maintaining localization performance. In addition, we also introduced a new sequence chunking approach that enables us to process very large scale datasets (up to 89,000 frames). We have empirically shown that our method achieves a better sub-linear compression ratio than~\cite{yu2018rhythmic} while also improving the querying time duration. The approach also exhibits improved robustness to varying environmental conditions through these changes and incorporation of a recently introduced feature LostX.

There are several promising avenues for future research. The chunking approach presented here is not necessarily limited to a single scale: it may be possible to perform chunking at several scales in a hierarchy, enabling further scalability: the key challenge here will be whether the chunk recognition schemes can still work reliably in a hierarchy. Another potential improvement could be achieved by replacing the linear pattern recognition and phase-encoding SVMs with deep neural networks using an end-to-end training framework: a further reduction in storage requirements would result if the deep net can learn a better representation for the phase encoders. Finally, our approach is generally agnostic of the application domain and can potentially be applied in other domains such as voice recognition and natural language processing to develop efficient encoding algorithms: further research in these additional domains will shed more light on the key relationship between the characteristics of the domain data and the sub-linear storage performance that is achieved.

\bibliographystyle{IEEEtran}
\bibliography{sublinear.bib}

\begin{thebibliography}{10}
\providecommand{\url}[1]{#1}
\csname url@samestyle\endcsname
\providecommand{\newblock}{\relax}
\providecommand{\bibinfo}[2]{#2}
\providecommand{\BIBentrySTDinterwordspacing}{\spaceskip=0pt\relax}
\providecommand{\BIBentryALTinterwordstretchfactor}{4}
\providecommand{\BIBentryALTinterwordspacing}{\spaceskip=\fontdimen2\font plus
\BIBentryALTinterwordstretchfactor\fontdimen3\font minus
  \fontdimen4\font\relax}
\providecommand{\BIBforeignlanguage}[2]{{%
\expandafter\ifx\csname l@#1\endcsname\relax
\typeout{** WARNING: IEEEtran.bst: No hyphenation pattern has been}%
\typeout{** loaded for the language `#1'. Using the pattern for}%
\typeout{** the default language instead.}%
\else
\language=\csname l@#1\endcsname
\fi
#2}}
\providecommand{\BIBdecl}{\relax}
\BIBdecl

\bibitem{yu2018rhythmic}
L.~Yu, A.~Jacobson, and M.~Milford, ``Rhythmic representations: Learning
  periodic patterns for scalable place recognition at a sublinear storage
  cost,'' \emph{IEEE Robotics and Automation Letters}, vol.~3, no.~2, pp.
  811--818, 2018.

\bibitem{lowry2016visual}
S.~Lowry, N.~S{\"u}nderhauf, P.~Newman, J.~J. Leonard, D.~Cox, P.~Corke, and
  M.~J. Milford, ``Visual place recognition: A survey,'' \emph{IEEE
  Transactions on Robotics}, vol.~32, no.~1, pp. 1--19, 2016.

\bibitem{lynen2015get}
S.~Lynen, T.~Sattler, M.~Bosse, J.~A. Hesch, M.~Pollefeys, and R.~Siegwart,
  ``Get out of my lab: Large-scale, real-time visual-inertial localization.''

\bibitem{cadena2016past}
C.~Cadena, L.~Carlone, H.~Carrillo, Y.~Latif, D.~Scaramuzza, J.~Neira, I.~Reid,
  and J.~J. Leonard, ``Past, present, and future of simultaneous localization
  and mapping: Toward the robust-perception age,'' \emph{IEEE Transactions on
  Robotics}, vol.~32, no.~6, pp. 1309--1332, 2016.

\bibitem{bojarski2016end}
M.~Bojarski, D.~Del~Testa, D.~Dworakowski, B.~Firner, B.~Flepp, P.~Goyal, L.~D.
  Jackel, M.~Monfort, U.~Muller, J.~Zhang \emph{et~al.}, ``End to end learning
  for self-driving cars,'' \emph{arXiv preprint arXiv:1604.07316}, 2016.

\bibitem{fyhn2004spatial}
M.~Fyhn, S.~Molden, M.~P. Witter, E.~I. Moser, and M.-B. Moser, ``Spatial
  representation in the entorhinal cortex,'' \emph{Science}, vol. 305, no.
  5688, pp. 1258--1264, 2004.

\bibitem{burak2009accurate}
Y.~Burak and I.~R. Fiete, ``Accurate path integration in continuous attractor
  network models of grid cells,'' \emph{PLoS computational biology}, vol.~5,
  no.~2, p. e1000291, 2009.

\bibitem{sreenivasan2011grid}
S.~Sreenivasan and I.~Fiete, ``Grid cells generate an analog error-correcting
  code for singularly precise neural computation,'' \emph{Nature neuroscience},
  vol.~14, no.~10, p. 1330, 2011.

\bibitem{garg2018lost}
S.~Garg, N.~Suenderhauf, and M.~Milford, ``Lost? appearance-invariant place
  recognition for opposite viewpoints using visual semantics,''
  \emph{Proceedings of Robotics: Science and Systems XIV}, 2018.

\bibitem{filliat2007visual}
D.~Filliat, ``A visual bag of words method for interactive qualitative
  localization and mapping,'' in \emph{Robotics and Automation, 2007 IEEE
  International Conference on}.\hskip 1em plus 0.5em minus 0.4em\relax IEEE,
  2007, pp. 3921--3926.

\bibitem{sivic2003video}
J.~Sivic and A.~Zisserman, ``Video google: A text retrieval approach to object
  matching in videos,'' in \emph{null}.\hskip 1em plus 0.5em minus 0.4em\relax
  IEEE, 2003, p. 1470.

\bibitem{nister2006scalable}
D.~Nister and H.~Stewenius, ``Scalable recognition with a vocabulary tree,'' in
  \emph{Computer vision and pattern recognition, 2006 IEEE computer society
  conference on}, vol.~2.\hskip 1em plus 0.5em minus 0.4em\relax Ieee, 2006,
  pp. 2161--2168.

\bibitem{gersho2012vector}
A.~Gersho and R.~M. Gray, \emph{Vector quantization and signal
  compression}.\hskip 1em plus 0.5em minus 0.4em\relax Springer Science \&
  Business Media, 2012, vol. 159.

\bibitem{jegou2011product}
H.~Jegou, M.~Douze, and C.~Schmid, ``Product quantization for nearest neighbor
  search,'' \emph{IEEE transactions on pattern analysis and machine
  intelligence}, vol.~33, no.~1, pp. 117--128, 2011.

\bibitem{kalantidis2014locally}
Y.~Kalantidis and Y.~Avrithis, ``Locally optimized product quantization for
  approximate nearest neighbor search,'' in \emph{Proceedings of the IEEE
  Conference on Computer Vision and Pattern Recognition}, 2014, pp. 2321--2328.

\bibitem{macqueen1967some}
J.~MacQueen \emph{et~al.}, ``Some methods for classification and analysis of
  multivariate observations,'' in \emph{Proceedings of the fifth Berkeley
  symposium on mathematical statistics and probability}, vol.~1, no.~14.\hskip
  1em plus 0.5em minus 0.4em\relax Oakland, CA, USA, 1967, pp. 281--297.

\bibitem{jegou2012aggregating}
H.~Jegou, F.~Perronnin, M.~Douze, J.~S{\'a}nchez, P.~Perez, and C.~Schmid,
  ``Aggregating local image descriptors into compact codes,'' \emph{IEEE
  transactions on pattern analysis and machine intelligence}, vol.~34, no.~9,
  pp. 1704--1716, 2012.

\bibitem{jegou2010aggregating}
H.~J{\'e}gou, M.~Douze, C.~Schmid, and P.~P{\'e}rez, ``Aggregating local
  descriptors into a compact image representation,'' in \emph{Computer Vision
  and Pattern Recognition (CVPR), 2010 IEEE Conference on}.\hskip 1em plus
  0.5em minus 0.4em\relax IEEE, 2010, pp. 3304--3311.

\bibitem{wang2018survey}
J.~Wang, T.~Zhang, N.~Sebe, H.~T. Shen \emph{et~al.}, ``A survey on learning to
  hash,'' \emph{IEEE transactions on pattern analysis and machine
  intelligence}, vol.~40, no.~4, pp. 769--790, 2018.

\bibitem{gong2013iterative}
Y.~Gong, S.~Lazebnik, A.~Gordo, and F.~Perronnin, ``Iterative quantization: A
  procrustean approach to learning binary codes for large-scale image
  retrieval,'' \emph{IEEE Transactions on Pattern Analysis and Machine
  Intelligence}, vol.~35, no.~12, pp. 2916--2929, 2013.

\bibitem{lin2015deep}
K.~Lin, H.-F. Yang, J.-H. Hsiao, and C.-S. Chen, ``Deep learning of binary hash
  codes for fast image retrieval,'' in \emph{Proceedings of the IEEE conference
  on computer vision and pattern recognition workshops}, 2015, pp. 27--35.

\bibitem{liu2016deep}
H.~Liu, R.~Wang, S.~Shan, and X.~Chen, ``Deep supervised hashing for fast image
  retrieval,'' in \emph{Proceedings of the IEEE conference on computer vision
  and pattern recognition}, 2016, pp. 2064--2072.

\bibitem{he2013k}
K.~He, F.~Wen, and J.~Sun, ``K-means hashing: An affinity-preserving
  quantization method for learning binary compact codes,'' in \emph{Proceedings
  of the IEEE conference on computer vision and pattern recognition}, 2013, pp.
  2938--2945.

\bibitem{giocomo2007temporal}
L.~M. Giocomo, E.~A. Zilli, E.~Frans{\'e}n, and M.~E. Hasselmo, ``Temporal
  frequency of subthreshold oscillations scales with entorhinal grid cell field
  spacing,'' \emph{Science}, vol. 315, no. 5819, pp. 1719--1722, 2007.

\bibitem{witten2010framework}
D.~M. Witten and R.~Tibshirani, ``A framework for feature selection in
  clustering,'' \emph{Journal of the American Statistical Association}, vol.
  105, no. 490, pp. 713--726, 2010.

\bibitem{ge2013optimized}
T.~Ge, K.~He, Q.~Ke, and J.~Sun, ``Optimized product quantization for
  approximate nearest neighbor search,'' in \emph{Computer Vision and Pattern
  Recognition (CVPR), 2013 IEEE Conference on}.\hskip 1em plus 0.5em minus
  0.4em\relax IEEE, 2013, pp. 2946--2953.

\bibitem{fan2008liblinear}
R.-E. Fan, K.-W. Chang, C.-J. Hsieh, X.-R. Wang, and C.-J. Lin, ``Liblinear: A
  library for large linear classification,'' \emph{Journal of machine learning
  research}, vol.~9, no. Aug, pp. 1871--1874, 2008.

\end{thebibliography}

\end{document}